\documentclass[
]{ceurart}

\usepackage{listings}
\usepackage{url}
\usepackage{hyperref}
\begin{document}

\copyrightyear{2022}
\copyrightclause{Copyright for this paper by its authors.
  Use permitted under Creative Commons License Attribution 4.0
  International (CC BY 4.0).}

\conference{Forum for Information Retrieval Evaluation, December 12-15, 2024, India}

\title{Cancer-Answer: Empowering Cancer Care with Advanced Large Language Models}


\author[1]{Aniket Deroy}[%
orcid=0000-0001-7190-5040,
email=roydanik18@kgpian.iitkgp.ac.in,
]
\cormark[1]
\fnmark[1]
\address[1]{IIT Kharagpur,
  Kharagpur, India}

\author[1]{Subhankar Maity}[%
orcid=0009-0001-1358-9534,
email=subhankar.ai@kgpian.iitkgp.ac.in,
]

\cortext[1]{Corresponding author.}

\begin{abstract}
Gastrointestinal (GI) tract cancers account for a substantial portion of the global cancer burden, where
early diagnosis is critical for improved management and patient outcomes. The complex aetiologies and
overlapping symptoms across GI cancers often delay diagnosis, leading to suboptimal treatment strategies.
Cancer-related queries are crucial for timely diagnosis, treatment, and patient education, as access to
accurate, comprehensive information can significantly influence outcomes. However, the complexity of
cancer as a disease, combined with the vast amount of available data, makes it difficult for clinicians and
patients to quickly find precise answers. To address these challenges, we leverage large language models
(LLMs) such as GPT-3.5 Turbo to generate accurate, contextually relevant responses to cancer-related
queries. Pre-trained with medical data, these models provide timely, actionable insights that support
informed decision-making in cancer diagnosis and care, ultimately improving patient outcomes. We calculate two metrics: A1 (which represents the fraction of entities present in the model-generated answer compared to the gold standard) and A2 (which represents the linguistic correctness and meaningfulness of the model-generated answer with respect to the gold standard), achieving maximum values of 0.546 and 0.881, respectively.

\end{abstract}


\maketitle

\section{Introduction}
Gastrointestinal (GI) tract cancers~\cite{bernstein2008field,bijlsma2017molecular,islami2004epidemiologic}, represent a significant portion of cancer-related morbidity and mortality worldwide, encompassing malignancies of the esophagus, stomach, liver, pancreas, and intestines. Early detection and accurate diagnosis are paramount for improving prognosis and patient survival. However, these cancers present a unique set of challenges due to their complex aetiologies and overlapping symptoms, which often result in delayed diagnosis and misclassification. Differentiating between various GI tract cancers remains a formidable task for clinicians, who must navigate a wide array of symptoms that can mimic benign conditions or other malignancies. In this context, cancer-related queries
are crucial for timely diagnosis, treatment, and patient education, as access to accurate and
comprehensive information can significantly impact outcomes \cite{c1}. The sheer volume of data and
complexity of cancer as a disease make it difficult for clinicians and patients to quickly access
precise answers.

Traditional diagnostic methods~\cite{rodriguez2006application,coda2014state}, including imaging, endoscopy, and histopathological examination, although valuable, sometimes fall short in providing rapid and precise differentiation of these cancer types. As a result, delays in diagnosis can compromise the effectiveness of treatment, leading to suboptimal patient outcomes. The need for more advanced, data-driven diagnostic tools has never been greater. In recent years, the advent of large language models (LLMs)~\cite{wilhelm2023large,nazi2024large} such as GPT-3.5 Turbo has opened new possibilities in medical diagnostics and decision support. These models, when prompted appropriately, have demonstrated remarkable potential in generating human-like text and answering complex queries. Leveraging LLMs for medical diagnostics offers a promising approach to addressing the diagnostic challenges posed by GI tract cancers. In this work, we explore the use of prompted LLMs to generate answers to medical queries, with a particular focus on their applicability to differentiating between GI tract cancers. By harnessing the power of these models, we aim to offer new insights into how artificial intelligence can assist clinicians in making more timely and accurate diagnoses, ultimately improving patient outcomes.

We explore prompted large language models (LLMs), such as GPT-3.5 Turbo ~\cite{brown2020language}, by designing and using specific prompts to generate relevant and coherent answers for various medical queries. These prompts guide the model to focus on producing medically accurate, context-appropriate responses based on the input. By leveraging the capabilities of GPT-3.5 Turbo, we aim to harness its vast knowledge base and advanced natural language understanding to assist in addressing a range of medical-related questions effectively. The use of prompts ensures that the model responds in a structured manner, providing meaningful information that aligns with the specific medical context of the queries.

Metrices A1 and A2 represent distinct components or processes within the evaluated system, and their performance metrics offer valuable information about the operational efficiency and potential areas for enhancement. By examining the results from multiple runs, we aim to identify trends, improvements, or inconsistencies that could impact the overall effectiveness of the system. This evaluation not only helps in understanding the current performance but also guides future adjustments and optimizations.

\section{Related Work}

The application of artificial intelligence (AI) in healthcare~\cite{shaheen2021applications,vaananen2021ai,panesar2019machine} has seen rapid growth in recent years, particularly in the areas of diagnostics and decision support. Several studies have explored the potential of machine learning (ML) and deep learning techniques to assist in the detection and classification of gastrointestinal (GI) cancers~\cite{rocken2017molecular,kuntz2021gastrointestinal,serra2019comparison}. These approaches range from traditional supervised learning models to more advanced AI systems, including convolutional neural networks (CNNs) and natural language processing (NLP) models.

Early work in AI-assisted GI cancer diagnostics primarily focused on image-based techniques. For instance, CNNs have been employed to analyze endoscopic and radiological images for detecting specific types of GI cancers such as esophageal and colorectal cancer~\cite{liu2020automatic,xie2022diagnostic,mohan2020accuracy}. Studies like ~\cite{mitsala2021artificial,he2021deep,omar2021deep} demonstrated the ability of deep learning algorithms to match or even surpass human-level performance in identifying cancerous lesions. While these methods have made significant strides, they are primarily limited to image processing tasks, requiring large amounts of annotated data and sophisticated preprocessing techniques.

NLP models, on the other hand, have recently been leveraged to analyze clinical reports and patient records. Various works, such as by ~\cite{zhang2020combining,spasic2020clinical,seinen2022use} utilized deep learning models to extract relevant information from unstructured text data, aiming to support clinical decision-making. However, these approaches often rely on predefined rules or training with vast, labeled datasets, which limits their generalizability to diverse clinical scenarios, including GI cancers.

The emergence of large language models (LLMs) like GPT-3 and GPT-3.5 has introduced a new frontier in NLP for healthcare~\cite{liu2023summary,hadi2023survey,hadi2023large}. These models are pre-trained on massive amounts of text data and can generate highly contextualized responses to medical queries with minimal fine-tuning. Studies such ~\cite{chen2024survey,khan2024comprehensive,treder2024introduction} have begun exploring the utility of LLMs in medical applications, showing promising results in generating accurate, coherent responses to clinical questions. However, most research to date has focused on general medical knowledge or specific diseases, with limited exploration into their potential for diagnosing GI tract cancers.

The idea of using prompted LLMs to aid in GI cancer diagnostics~\cite{liu2023summary,hager2024evaluating,hager2024evaluation} remains underexplored. While existing NLP systems provide valuable insights, they often lack the ability to dynamically respond to complex medical queries without extensive fine-tuning. In contrast, GPT-3.5 Turbo and similar models can be prompted to generate medically relevant text with minimal training, potentially addressing some of the limitations faced by previous systems. This work builds on the foundation of AI in healthcare by investigating the use of prompted LLMs to generate responses to GI cancer-related medical queries, contributing to the growing body of research on AI-powered diagnostics in oncology.
\section{Dataset}
There are 30 queries in the training set related to GI-Cancer. There are 50 queries in the testing set related to GI-Cancer.

\section{Task Definition}
We need to design a Question Answering based conversational system  that can provide answers to queries related to GI Cancer, using an AI model.

\section{Methodology}

Prompting \cite{c2} is an emerging technique in the development of question-answering (QA) systems, particularly with the advent of large language models (LLMs) like GPT-3.5. Unlike traditional machine learning methods, which often require large amounts of labeled data and extensive fine-tuning, prompting involves crafting specific input instructions that guide LLMs to generate relevant answers to user queries. This approach has been tried for these key reasons:

\begin{itemize}[-]
   
\item \textbf{Minimal Data and Fine-Tuning Requirements:} One of the major advantages of prompting is that it minimizes the need for extensive training data and domain-specific fine-tuning~\cite{ge2023domain}. Traditional QA systems rely on massive datasets to train models for each specific task. With prompting, models like GPT-3.5 can leverage their pre-existing knowledge from vast amounts of pre-trained data, allowing them to generate accurate answers across different domains with minimal additional data. This is particularly useful in medical domains like gastrointestinal (GI) cancer diagnostics, where high-quality labeled data can be scarce and time-consuming to acquire.

\item \textbf{Generalization Across Diverse Topics:} LLMs pre-trained on large and diverse corpora are capable of handling a wide variety of questions without being confined to a narrow domain~\cite{patil2024review}. In contrast, conventional QA systems typically require specialized models for specific areas. By using well-designed prompts, LLMs can provide answers to medical questions across different GI tract cancers without needing separate models for each type of cancer or medical condition. This flexibility allows the same model to respond to queries about symptoms, diagnostics, and treatment, improving efficiency.

\item \textbf{Reduced Development Time:} The prompt-based approach reduces the time and complexity involved in developing QA systems~\cite{gekhman2023robustness}. Traditional systems require careful preprocessing, feature extraction, and extensive model training. By contrast, prompting requires only well-constructed input prompts that instruct the LLM to generate a response. This simplifies the development process and allows for rapid iteration, enabling QA systems to be deployed quickly in clinical environments.

\item \textbf{Dynamic and Contextual Responses:} LLMs are designed to understand the context of a question and generate dynamic, human-like responses~\cite{alsafari2024towards}. By using specific prompts, QA systems can better interpret the nuances of medical questions, which is critical in complex domains such as GI cancer. The models can adapt to variations in question phrasing, offering contextually relevant answers that align with the complexity of medical knowledge. For example, they can handle follow-up questions or clarify answers based on additional information provided by the user.

\item \textbf{Scalability and Adaptability:} Prompt-based QA systems are highly scalable~\cite{alsafari2024towards}, as they do not require retraining or large-scale infrastructure changes when applied to new domains or updated with new information. This is particularly useful in rapidly evolving fields like medicine, where new research and findings continuously emerge. The adaptability of LLMs to new topics through updated prompts allows QA systems to stay current with the latest medical knowledge without the need for re-engineering the entire system.

\item \textbf{Cost-Effective Solution:} Developing and maintaining traditional QA systems can be resource-intensive due to the need for large datasets, computing power, and expertise in model training~\cite{cohen2021methodology}. Prompting offers a cost-effective alternative, as it capitalizes on the power of pre-trained LLMs. This approach reduces the dependency on large-scale infrastructure and can be easily implemented without requiring extensive computational resources.
 
\end{itemize}
In summary, prompting is an efficient, flexible, and scalable solution for building question-answering systems in specialized domains like medical diagnostics. By leveraging LLMs through well-designed prompts, QA systems can generate accurate, context-aware responses, significantly reducing development time, data requirements, and costs, while improving the overall quality and accessibility of information. For the field of GI cancer diagnostics, prompted LLMs offer a promising tool for clinicians, allowing them to access critical information and make informed decisions more effectively.

\subsection{Prompt Engineering-Based Approach}
We used the GPT-3.5 Turbo\footnote{\url{https://platform.openai.com/docs/models/gpt-3-5-turbo}} model via prompting to solve the question-answering task.
We used GPT-3.5 Turbo in zero-Shot mode via prompting.
After the prompt is provided to the LLM, the following steps happen internal to the LLM while generating the output. The following outlines the steps that occur internally within the LLM, summarizing the prompting approach using GPT-3.5 Turbo:\\

\textbf{Step 1: Tokenization}

\begin{itemize}
    \item \textbf{Prompt:} \( X = [x_1, x_2, \dots, x_n] \)
    \item The input text (prompt) is first tokenized into smaller units called tokens. These tokens are often subwords or characters, depending on the model's design.
    \item \textbf{Tokenized Input:} \( T = [t_1, t_2, \dots, t_m] \)
\end{itemize}

\textbf{Step 2: Embedding}

\begin{itemize}
    \item Each token is converted into a high-dimensional vector (embedding) using an embedding matrix \( E \).
    \item \textbf{Embedding Matrix:} \( E \in \mathbb{R}^{|V| \times d} \), where \( |V| \) is the size of the vocabulary and \( d \) is the embedding dimension.
    \item \textbf{Embedded Tokens:} \( T_{\text{emb}} = [E(t_1), E(t_2), \dots, E(t_m)] \)
\end{itemize}

\textbf{Step 3: Positional Encoding}

\begin{itemize}
    \item Since the model processes sequences, it adds positional information to the embeddings to capture the order of tokens.
    \item \textbf{Positional Encoding:} \( P(t_i) \)
    \item \textbf{Input to the Model:} \( Z = T_{\text{emb}} + P \)
\end{itemize}

\textbf{Step 4: Attention Mechanism (Transformer Architecture)}

\begin{itemize}
    \item \textbf{Attention Score Calculation:} The model computes attention scores to determine the importance of each token relative to others in the sequence.
    \item \textbf{Attention Formula:}
    \begin{equation}
    \text{Attention}(Q, K, V) = \text{softmax}\left(\frac{QK^T}{\sqrt{d_k}}\right)V
    \end{equation}
    \item where \( Q \) (query), \( K \) (key), and \( V \) (value) are linear transformations of the input \( Z \).
    \item This attention mechanism is applied multiple times through multi-head attention, allowing the model to focus on different parts of the sequence simultaneously.
\end{itemize}

\textbf{Step 5: Feedforward Neural Networks}

\begin{itemize}
    \item The output of the attention mechanism is passed through feedforward neural networks, which apply non-linear transformations.
    \item \textbf{Feedforward Layer:}
    \begin{equation}
    \text{FFN}(x) = \max(0, xW_1 + b_1)W_2 + b_2
    \end{equation}
    \item where \( W_1, W_2 \) are weight matrices and \( b_1, b_2 \) are biases.
\end{itemize}

\textbf{Step 6: Stacking Layers}

\begin{itemize}
    \item Multiple layers of attention and feedforward networks are stacked, each with its own set of parameters. This forms the "deep" in deep learning.
    \item \textbf{Layer Output:}
    \begin{equation}
    H^{(l)} = \text{LayerNorm}(Z^{(l)} + \text{Attention}(Q^{(l)}, K^{(l)}, V^{(l)}))
    \end{equation}
    \begin{equation}
    Z^{(l+1)} = \text{LayerNorm}(H^{(l)} + \text{FFN}(H^{(l)}))
    \end{equation}
\end{itemize}

\textbf{Step 7: Output Generation}

\begin{itemize}
    \item The final output of the stacked layers is a sequence of vectors.
    \item These vectors are projected back into the token space using a softmax layer to predict the next token or word in the sequence.
    \item \textbf{Softmax Function:}
    \begin{equation}
    P(y_i|X) = \frac{\exp(Z_i)}{\sum_{j=1}^{|V|} \exp(Z_j)}
    \end{equation}
    \item where \( Z_i \) is the logit corresponding to token \( i \) in the vocabulary.
    \item The model generates the next token in the sequence based on the probability distribution, and the process repeats until the end of the output sequence is reached.
\end{itemize}

\textbf{Step 8: Decoding}

\begin{itemize}
    \item The predicted tokens are then decoded back into text, forming the final output.
    \item \textbf{Output Text:} \( Y = [y_1, y_2, \dots, y_k] \)
\end{itemize}

\begin{figure}[h!]
  \centering
  \includegraphics[width=0.80\linewidth]{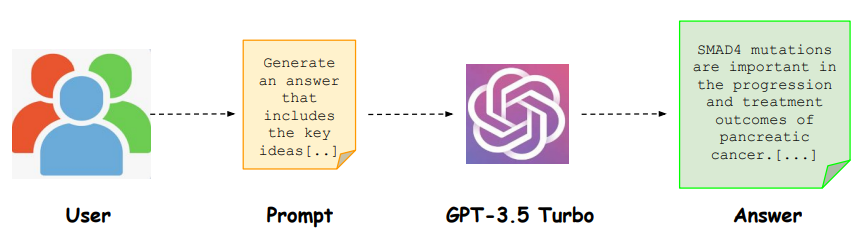}
  \caption{Flow diagram depicting the interaction between a user and GPT-3.5 Turbo for answering cancer-related queries. The diagram details the steps from user prompt submission to the generation of responses/answers by the LLM.} \label{fig1}
\end{figure}

\begin{figure}[h!]
  \centering
  \includegraphics[width=0.60\linewidth]{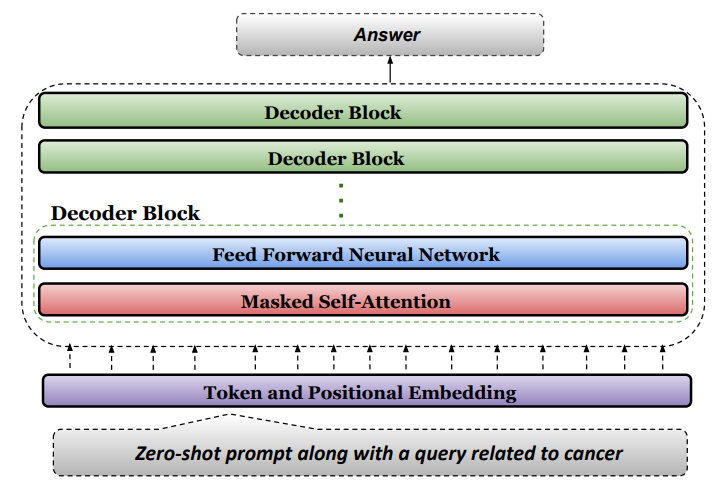}
  \caption{An Overview of GPT-3.5 Turbo for Answering Cancer-Related Queries.} \label{fig2}
\end{figure}

A flow diagram depicting the interaction between a user and GPT-3.5 Turbo to answer cancer-related queries is shown in Figure ~\ref{fig1}. An overview of GPT-3.5 Turbo to answer cancer-related queries is shown in Figure ~\ref{fig2}.

We use three different prompts to GPT-3.5 Turbo as follows:
\begin{enumerate}[(i)]
    
\item We used GPT-3.5 Turbo in zero-shot mode at temperature 0.7 with the following prompt:
"\textit{{Generate an answer that includes the key ideas corresponding to the question <question>}.}"

\item We used GPT-3.5 Turbo in zero-shot mode at temperature 0.7 with a simpler prompt:\textit{ "{<query>}"}
(The query was directly passed as the prompt without additional instructions.)

\item We used GPT-3.5 Turbo in zero-shot mode at temperature 0.7 with the following prompt: \textit{"Please summarize the key ideas of the answer to the following cancer-related question in one paragraph: <question>"}.

\end{enumerate}

\section{Results}

We have defined two metrics namely:

\textbf{A1}: A1 measures the fraction of entities present in the model-generated answers which are also present in the gold standard answers.

\[
\text{A1} = \frac{|\text{(Entities in Model Generated Answer)} \cap \text{(Entities in Gold Standard Answer)}|}{|\text{(Entities in Gold Standard Answer)}|}
\]

We manually calculate the entities for every model-generated answer and the gold standard answer to calculate the value of A1 for a particular (Query, Answer) pair. The values are added over the 50 samples in the test set and then averaged out of the 50 samples.

\textbf{A2}: A2 measures the linguistic correctness and meaningfulness of the model-generated answers wrt to the gold standard data.

We manually assign a value of [0,1] representing the linguistic correctness and meaningfulness of the generated answer to calculate the value of A2 for a particular (Query, Answer) pair. The values are added over the 50 samples in the test set and then averaged out of the 50 samples.


Table~\ref{tab:values} shows the values for three different runs for the two metrics, namely A1, and A2.

For A1, A2, the upward trend is positive and suggests that the metric being measured is becoming more effective or efficient. It might be beneficial to investigate what changes were made between runs that led to these improvements.


\begin{table}[h!]
\centering
\begin{tabular}{lccc}
\toprule
       & \textbf{Run 1} & \textbf{Run 2} & \textbf{Run 3} \\
\midrule
\textbf{A1} & 0.354 & 0.452 & 0.546 \\
\textbf{A2} & 0.622 & 0.702 & 0.881 \\
\bottomrule
\end{tabular}
\caption{Values for A1 and A2 across three runs.}
\label{tab:values}
\end{table}




\section{Conclusion}
In this work, we explored the potential of large language models (LLMs), particularly GPT-3.5 Turbo, as a question-answering (QA) tool to address the challenges associated with diagnosing gastrointestinal (GI) tract cancers. GI cancers pose unique difficulties due to overlapping symptoms and complex aetiologies, often leading to delayed diagnoses and suboptimal treatment strategies. By leveraging the power of prompted LLMs, we demonstrated the capability of these models to generate coherent, contextually relevant answers to medical queries, providing a flexible and efficient approach for assisting clinicians in differentiating between various GI cancers.

The analysis of performance metrics for metrics A1 and A2 across three runs reveals important insights into their behavior and effectiveness. For metric A1, A2 the consistent improvement in performance across the runs indicates a successful enhancement of the underlying system or methodology. This positive trend suggests that the adjustments or optimizations implemented are yielding favorable results and warrants continued focus and refinement.


Our findings highlight the advantages of using prompt-based systems in healthcare, including the ability to generalize across a wide range of medical topics, minimal data requirements, and the flexibility to dynamically adapt to new information. These characteristics make LLMs a promising tool for augmenting clinical decision-making, particularly in resource-constrained environments where access to specialized diagnostic expertise may be limited.

However, it is important to recognize the limitations of current LLMs in handling highly specialized or nuanced medical cases, underscoring the need for ongoing research to improve model accuracy and reliability. Future work could focus on further fine-tuning LLMs with domain-specific data or incorporating additional knowledge sources to enhance their diagnostic capabilities.

Overall, the integration of LLMs into clinical workflows has the potential to improve the accuracy and timeliness of cancer diagnoses, particularly in complex cases like GI tract cancers, ultimately contributing to better patient outcomes and more efficient healthcare delivery.


\bibliography{sample-ceur}

\end{document}